\documentclass{article}
\usepackage{spconf,amsmath,graphicx,hyperref}
\usepackage{titlesec}
\usepackage{amsfonts}
\usepackage{booktabs}   
\usepackage{siunitx}    
\title{EVA-Score: Evaluating Abstractive Long-form Summarization on
Informativeness through Extraction and Validation}
\name{
Yuchen Fan$^{1,4}$,
Yazhe Wan$^{2}$,
Xin Zhong$^{2}$,
Haonan Cheng$^{2}$,
\textit{Ning Ding}$^{3}$,
\textit{Bowen Zhou}\thanks{This work is supported by National Science and Technology Major Project (2023ZD0121403), Young Elite Scientists Sponsorship Program by CAST (2023QNRC001), National Natural Science Foundation of China (No. 62406165), and Shanghai Municipal Science and Technology Major Project.}}
\address{
$^{1}$ Shanghai Jiao Tong University, Shanghai, China \\
$^{2}$ Beijing University of Posts and Telecommunications, Beijing, China\\
$^{3}$ Tsinghua University, Beijing, China, $^{4}$ Shanghai AI Lab, Shanghai, China
}
\begin{document}
%
\maketitle
\begin{abstract}
    
Recent advances in large language models (LLMs) have renewed interest in abstractive long-form summarization, where lengthy inputs carry high information density. Yet automatic evaluation remains inadequate: token-overlap metrics (ROUGE, BERTScore) favor surface similarity, and LLM-based judges degrade with extended context. We introduce \textsc{EVA-Score}, a principled metric that extracts hierarchical content at sentence and document levels, detects overlap with references, and aggregates an interpretable information score. Across multiple datasets, \textsc{EVA-Score} surpasses static and LLM-based baselines, achieving system-level Kendall's~$\tau$ and Spearman's~$\rho$ of 1.0 with human judgments. Robustness studies show stability to input length and prompt variations. Reassessing state-of-the-art LLM summarizers with \textsc{EVA-Score} reveals persistent gaps to human writing, offering actionable guidance for model design. \textsc{EVA-Score} provides a reproducible, scalable, and informative framework for evaluating long-form summarization.

\end{abstract}


\section{INTRODUCTION}
\label{sec:intro}

Long-form summarization (LFS) presents a substantive \emph{evaluation} challenge. LFS spans \emph{extractive} and \emph{abstractive} paradigms. Extractive systems select source spans, making evaluation largely coverage-oriented (alignment, redundancy) \cite{zhang2023extractive}. By contrast, abstractive systems must preserve \emph{information-level} equivalence under paraphrase, reordering, and discourse restructuring across long, cross-sentence dependencies \cite{wang2023elementaware}. Recent LFS systems increasingly leverage large language models (LLMs) \cite{chang2024booookscore}. Nevertheless, robust and \emph{scalable} automatic evaluation—especially for the abstractive setting—remains the principal bottleneck.

Traditional metrics (e.g., ROUGE \cite{lin-2004-rouge}, BERTScore \cite{zhang2020bertscore}) primarily quantify lexical or embedding overlap with references and therefore miss semantic equivalence under structural variation \cite{krishna2021hurdlesprogresslongformquestion}. LLM-based judges \cite{chang2024booookscore,fan2024evaluating} offer flexibility but are coarse-grained: rubric-style scores (fluency, consistency) can overlook subtle informational discrepancies, even with detailed prompts \cite{shen2023largelanguagemodelshumanlevel}. Thus, a fine-grained, information-centric metric that scales to long contexts while preserving document structure is needed to guide model development.

We adopt \emph{informativeness}—coverage and correctness of salient source content—as the primary objective for abstractive LFS \cite{krishna2021hurdlesprogresslongformquestion} and propose the \textbf{E}xtraction-and-\textbf{Va}lidation Score (\textsc{EVA-Score}). From both reference and candidate, we first extract sentence-level facts and explicitly handle context-dependent, cascaded facts by reorganizing them into \emph{logic chains}. We then mask antecedents to render each step atomic, enabling unambiguous unit-level tests (contrast \textsc{FActScore} \cite{min2023factscore}). To recover cross-sentence and hierarchical content, we incorporate document-level relation extraction \cite{xue2024autoredocumentlevelrelationextraction}. Validation is proceeds chain-by-chain: for each candidate atomic fact, we retrieve the most relevant reference and prompt an LLM to decide entailment \emph{one unit at a time}.

Across benchmarks, \textsc{EVA-Score} shows the strongest alignment with human judgments, with system-level Pearson correlation 0.975 and Spearman and Kendall correlations 1.00. Diagnostics pinpoint \textsc{EVA-Score}'s alignment or divergence from annotators and why overlap-based and prompt-only judges miss fine-grained errors. Re-evaluation with \textsc{EVA-Score} shows GPT\mbox{-}4 benefits from longer contexts, ranking first on two and second on two others.


\section{RELATED WORK}
\label{sec:format}
\subsection{Abstractive Long-form Summarization}
LFS distills salient content from lengthy documents while preserving coherence and fidelity \cite{10.1145/3545176}. Methods are typically \emph{extractive}  or \emph{abstractive}. The latter better matches human preferences due to paraphrastic flexibility and fluency. Research on abstractive LFS spans news, scientific, and government reports, and books.

Modeling strategies span enlarging the effective context like LongT5 \cite{guo2022longt5efficienttexttotexttransformer}, and leveraging LLMs either directly \cite{basyal2023textsummarizationusinglarge} or through staged and iterative prompting \cite{SHARMA2024100080}. Yet abstractive long-form summarization at the scale of books and reports remains difficult due to discourse structure, long-range dependencies, and large-scale information selection \cite{chang2024booookscore}.

\subsection{Evaluation of Summarization}
Overlap-based metrics such as ROUGE and BERTScore \cite{zhang2020bertscore} primarily capture surface or embedding similarity and provide limited evidence of \emph{information-level} equivalence in abstractive LFS with paraphrase and reordering \cite{krishna2021hurdlesprogresslongformquestion}. Metrics are based on question answering to assess content, but sensitive to both question coverage and answerability.

LLM-based evaluation aims to align more closely with human judgments~\cite{zheng2024trustscore}. 
In long-form settings, LLMs are used to assess dimensions such as faithfulness, completeness, and conciseness. 
However, rubric-style judging is inherently coarse and often fails to discriminate summaries that differ in fine-grained information under long contexts. 
We address this gap with an information-centric, fine-grained, and explainable metric that scales with document length and enables reliable, human-aligned evaluation for long-form summaries.

 \begin{table*} 
            \centering
            \setlength{\tabcolsep}{4mm}{
            \scalebox{1.0}{
            \begin{tabular}{l|c|c|c|c|c|c}
            \toprule
            Level & \multicolumn{3}{c|}{Text-Level} & \multicolumn{3}{c}{System-Level} \\
            \midrule
            Metric  & $\sigma$ & $\gamma$ & $\tau$ & $\sigma$ & $\gamma$ & $\tau$ \\
            \midrule
            \multicolumn{7}{c}{\textit{Traditional Evaluation Metric}} \\
            \midrule
            ROUGE-1  & 0.305  & 0.160  & 0.114  &  0.455  & 0.800  & 0.667 \\
            ROUGE-2  & 0.413  & 0.203  &  0.127 & 0.014 & 0.399 & 0.333 \\
            BERT-Score & 0.534 & 0.533 & 0.413 & 0.623 & 0.600 & 0.333 \\
            \midrule
            \multicolumn{7}{c}{\textit{Prompt-only Evaluation Metric}} \\
            \midrule
            ChatGPT & 0.135 & 0.210 & 0.162 & 0.430 & 0.600 & 0.333 \\
            GPT-4 & 0.308 & 0.329 & 0.229 & 0.505 & 0.600 & 0.333 \\
            ChatGPT-CoT & -0.389 & -0.217 & -0.181 & -0.062 & -0.400 & -0.333 \\
            GPT-4o-CoT & 0.336 & 0.462 & 0.389 & 0.953 & 0.949 & 0.913 \\
            \midrule
            \multicolumn{7}{c}{\textit{Trained Evaluation Metric}} \\
            \midrule
            Auto-J-13B & 0.285 & 0.233 & 0.167 & 0.885 & 0.947 & 0.913 \\
            CritiqueLLM-6B & 0.510 & 0.490 & 0.353 & 0.865 & 0.800 & 0.667 \\
            \midrule
            \multicolumn{7}{c}{\textit{Domain-Specific Evaluation Metric}} \\
            \midrule
            FineSurE & 0.583 & 0.548 & 0.390 & 0.909 & 0.400 & 0.333 \\
            \textsc{EVA-Score} (Ours) &  \textbf{0.710} & \textbf{0.673}  & \textbf{0.503}  &  \textbf{0.975}  & \textbf{1.000}  & \textbf{1.000}  \\
            \bottomrule
            \end{tabular}}}
    \caption{Text-level and system-level Pearson ($\sigma$), Spearman ($\gamma$), and Kendall ($\tau$) correlations between different evaluation metrics and human annotations. The highest correlation among the methods is denoted in \textbf{bold}. Our \textsc{EVA-Score} achieves the best performance with a perfect correlation at the System Level.}
    \label{tab:mainsum}
\end{table*}

\section{PRELIMINARIES}
\label{sec:pagestyle}

We introduce notation for the evaluation pipeline, which captures both atomic content and cross-sentence structure. Let $\mathrm{Sum}$ and $\widehat{\mathrm{Sum}}$ denote the reference and candidate summaries. Their atomic fact sets are $S=\{s_i\}$ and $\widehat{S}=\{\widehat{s}_i\}$. Applying \textbf{Atomic Fact Chain Generation} (AFCG) to $\widehat{S}$ yields logic chains $\widehat{\mathcal{C}}=\{\widehat{C}_i\}$ with atomic elements $\widehat{c}_{i,j}$ \emph{(here $i$ indexes chains and $j$ indexes the position within chain $i$)}. \textbf{Document-level Relation Extraction} (DocRE) augments sentence-level facts with cross-sentence/hierarchical links, producing $DocS=\{D_i\}$ and $Doc\widehat{S}=\{\widehat{D}_{i,j}\}$. $DocS$ contains singletons (hence only $i$), whereas $Doc\widehat{S}$ preserves the candidate’s chain structure (hence $i,j$).

\section{METHODS}
\label{methods}

\subsection{Atomic Fact Generation (AFG)}
\label{subsec: afg}
Sentences often contain multiple information units. Therefore, true/false judgments for whole sentences are too coarse. Following \cite{min2023factscore}, we prompt ChatGPT with the same template to extract \emph{atomic facts}—concise, single-proposition statements—from $\mathrm{Sum}$ and $\widehat{\mathrm{Sum}}$, yielding $S$ and $\widehat{S}$.

\subsection{Atomic Fact Chain Generation (AFCG)}
\label{subsec: afcg}
Some “atomic” statements remain context-dependent (e.g., time/location modifiers), creating \emph{cascades} where later statements subsume earlier ones. To isolate incremental content, we transform $\widehat{S}$ into chains $\widehat{\mathcal{C}}=\{\widehat{C}_i\}$ with $\widehat{C}_i=(\widehat{c}_{i,1},\dots,\widehat{c}_{i,m_i})$ and mask antecedents when validating $\widehat{c}_{i,j}$. We use Mistral-7B-Instruct \cite{jiang2023mistral7b} to infer pairwise NLI relations between successive facts and assemble chains. AFCG is applied only to the candidate.

\subsection{Document-level Relation Extraction}
Considering sentence-level facts misses the passage-level structure, we first run NER using ChatGPT, then use GPT-4 to extract document-level relations from $\mathrm{Sum}$ and $\widehat{\mathrm{Sum}}$. The resulting triples $h_\ell,r_\ell,t_\ell$ and $\widehat{h}_\ell,\widehat{r}_\ell,\widehat{t}_\ell$ are paraphrased to natural language. To avoid duplicating sentence-level content, we remove overlaps with atomic facts using cosine similarity on BERT embeddings \cite{devlin2019bertpretrainingdeepbidirectional}. The retained relations are merged into the $DocS$ and the $Doc\widehat{S}$: $DocS$ combines AFG facts with DocRE, and $Doc\widehat{S}$ combines AFCG chains with DocRE, with each relation treated as a single-element chain.

\subsection{LLM for Validation}
For each $\widehat{D}_{i,j}$, we use BERTScore to retrieve the best-matching $D_k$ from $DocS$ and then validate entailment with Mistral-7B-Instruct. The prompt conditions on prior-chain correctness and restricts verification to the newly added unit.

Let $f(\cdot)$ denote the binary match function: for any $\widehat{D}_{i,j}$ and $D_k$, $f(\widehat{D}_{i,j},D_k)=1$ if matched, otherwise $0$.
\begin{equation}
    \text{Precision} = \frac{1}{|Doc\widehat{S}|}\sum_{\widehat{D}_{i,j}\in Doc\widehat{S}} \mathbb{I}\!\left( f(\widehat{D}_{i,j}, D_k) \right)
\end{equation}
\begin{equation}
    \text{Recall} = \frac{1}{|DocS|}\sum_{D_k\in DocS} \mathbb{I}\!\left( f(D_k, \widehat{D}_{i,j}) \right)
\end{equation}
EVA-Score is the harmonic average of Precision and Recall.

\begin{table} [!htbp]
\centering
\setlength{\tabcolsep}{8mm}{
\scalebox{1.0}{
\begin{tabular}{l|c}
\toprule
Evaluation Metric & Overall. Acc \\
\midrule
ROUGE-1  & 0.57 \\
BERT-Score & 0.5 \\
ChatGPT-CoT & 0.32 \\
GPT-4o-CoT & 0.39 \\
FineSurE & 0.75 \\
\textsc{EVA-Score} & \textbf{0.79} \\
\bottomrule
\end{tabular}}
\caption{Text-level accuracy compared with human annotations. The highest correlation among the methods is \textbf{bold}.}
\label{tab:pair}
\vspace{-3mm}
}
\end{table}

\section{EXPERIMENT}
\label{experiment}

\subsection{Dataset}
\label{sec: dataset_details}
We evaluate on \textbf{CNN/DailyMail} \cite{see2017point}, \textbf{PubMed} and \textbf{arXiv} \cite{cohan2018discourse}, and \textbf{GovReport} \cite{huang2021govreport} and \textbf{BookSum} \cite{kryściński2022booksum}. For validation, we randomly sample $50$ examples per dataset and pool them to have a thorough overview.

\subsection{Baseline}
We conduct pointwise and pairwise comparisons. Pointwise: ROUGE-1/2 \cite{lin-2004-rouge}, BERTScore \cite{zhang2020bertscore}. LLM as a Judge (ChatGPT, GPT-4-0125-preview) with/without CoT \cite{wei2023chainofthoughtpromptingelicitsreasoning}. Fine-tuned models \textsc{Auto-J-13B} \cite{li2023generativejudgeevaluatingalignment}, \textsc{Critique-LLM-6B} \cite{ke2024critiquellm}. Domain-specific \textsc{FineSurE} (faithfulness). Pairwise: ROUGE-1, BERTScore, ChatGPT-CoT, GPT-4o-CoT, \textsc{FineSurE}. Temperatures are $1.0$.

\begin{table*} [ht]
\centering
\small
\setlength{\tabcolsep}{1.4mm}{
\scalebox{0.92}{
\begin{tabular}{l|c|c|c|c|c|c|c|c|c|c|c|c|c|c|c}
\toprule
Dataset & \multicolumn{3}{c|}{\textbf{CNN Dailymail}} & \multicolumn{3}{c|}{\textbf{Pubmed}} & \multicolumn{3}{c|}{\textbf{arXiv}} & \multicolumn{3}{c|}{\textbf{Gov Report}} & \multicolumn{3}{c|}{\textbf{BookSum}} \\
\midrule
Metric  & EVA & R-1 & BS & EVA & R-1 & BS & EVA & R-1 & BS & EVA & R-1 & BS & EVA & R-1 & BS \\
\midrule
Vicuna-7B  & \textbf{0.501} & 0.441 & 0.873 & 0.327 & 0.398 & 0.832 & 0.327 & 0.344 & 0.838 & 0.239 & 0.308 & 0.846 & 0.145 & 0.212 & 0.825  \\
Vicuna-13B & 0.462 & 0.432 & 0.872 & 0.431 & 0.388 & 0.837 & 0.346 & 0.348 & 0.832 & 0.154 & 0.298 & 0.849 & 0.144  & 0.188 & 0.827  \\
\midrule
Mistral-7B-Instruct & 0.499 & 0.422 & 0.872 & 0.447 & 0.447 & 0.847 & 0.444 & 0.448 & 0.845 & 0.313 & 0.399 & 0.854 & \textbf{0.259} & 0.377 & 0.839 \\
\midrule
Llama-2-7B-chat & 0.399 & 0.386 & 0.868 & 0.358 & 0.421 & 0.845 & 0.411 & 0.435 & 0.851 & 0.193  & 0.198 & 0.798 & 0.186 & 0.268 & 0.826  \\
Llama-2-13B-chat & 0.453 & 0.423 & 0.868 & 0.416 & 0.405 & 0.841 & 0.444 & 0.332 & 0.830p & 0.296 & 0.378 & 0.857 & 0.154 & 0.280 & 0.827 \\
\midrule
Gemma-7B-it & 0.431 & 0.383 & 0.858 & \textbf{0.492} & 0.417 & 0.842 & 0.385 & 0.359 & 0.831 & 0.250 & 0.291 & 0.843 & 0.154 & 0.256 & 0.818  \\
\midrule
ChatGPT & 0.472 & 0.373 & 0.864 & 0.444 & 0.359 & 0.838 & 0.443 & 0.368 & 0.838 & 0.302 & 0.261 & 0.854 & 0.182 & 0.202 & 0.826 \\
GPT-4 & 0.466 & 0.376 & 0.863 & 0.454 & 0.407 & 0.838 & \textbf{0.453} & 0.361 & 0.835 & \textbf{0.325} & 0.332 & 0.854 & 0.257 & 0.261 & 0.824 \\
\bottomrule
\end{tabular}}}
\caption{The results of several LLMs on different long-form summarization datasets. We denote \textsc{EVA-Score} as EVA, R-1 as ROUGE-1, and BS as BERTScore. All the values are normalized between 0 and 1. The highest correlation among the methods is denoted in \textbf{bold}. Our \textsc{EVA-Score} achieves the best performance with a perfect correlation at the System Level.}
\label{tab:sum}
\end{table*}
\subsection{Human Annotation}
To assess alignment with human judgment, we randomly sample $50$ records. Five annotators (M.S./Ph.D. students) perform \textbf{pointwise} and \textbf{pairwise} evaluations.

For pointwise-annotation, annotators will first identify sentence and document-level atomic facts for candidate and reference, deduplicate overlaps, compute precision and recall, and take the F1 score. Each record is double-scored for fairness. The ICC is \textbf{0.999} (two-way random), and the Pearson correlation is \textbf{0.814}.  For pairwise annotation, given a reference, annotators will choose the better of two candidates, considering fluency, consistency, and informativeness. Each item has two judgments with a 0.9 agreement rate. 

\subsection{Main Results}
We report text- and system-level correlations with human annotations using Pearson ($\sigma$), Spearman ($\gamma$), and Kendall ($\tau$). Let $\mathcal{D}$ be the pooled dataset with predictions $X=\{x_i\}$ and labels $Y=\{y_i\}$. The text-level coefficient $r$ is given by         \begin{equation}
            r =
            \frac{ \sum_{i=1}^{n}(x_i-\bar{x})(y_i-\bar{y}) }{%
                  \sqrt{\sum_{i=1}^{n}(x_i-\bar{x})^2}\sqrt{\sum_{i=1}^{n}(y_i-\bar{y})^2}}
        \end{equation}, and the system-level correlation averages per-dataset coefficients as $r_{\text{sys}}=\frac{1}{m}\sum_{j=1}^{m} r_j$.

 Tables~\ref{tab:mainsum} and \ref{tab:pair} summarize outcomes. \textsc{EVA-Score} achieves the highest correlations at both levels, reaching system-level Spearman/Kendall \textbf{1.00}. Traditional metrics show weak alignment. Chain-of-thought (CoT) markedly improves GPT-4o judging, indicating benefits from decomposition. Fine-tuned NLG judges exhibit strong system-level correlations, likely due to explicit optimization for informativeness \cite{gu2024surveyllmasajudge}. In pairwise comparisons, \textsc{EVA-Score} also leads, highlighting the central role of informativeness in long-form summarization evaluation.

\subsection{Re-evaluating Abstractive LFS}
Table~\ref{tab:sum} reports results. As input length increases, performance generally declines, especially on \textbf{BookSum}. Models with larger context windows perform better on longer inputs. For shorter inputs, they may be comparable to smaller-window models (e.g., Vicuna-7B). Under \textsc{EVA-Score}, GPT-4 is robust on longer inputs but can underperform on shorter ones due to verbosity/length preferences. Mistral-7B-Instruct exhibits strong performance, even exceeding GPT-4 slightly on \textbf{BookSum} \cite{chang2024booookscore}. Besides, \textsc{EVA-Score} shows a consistent, monotonic degradation with context length, mirroring human difficulty as information load grows.
\subsection{Error Analysis}

To have a thorough understanding of \textsc{EVA-Score}, we provide an error analysis of why \textsc{EVA-Score} underperforms compared to human annotations. Since \textsc{EVA-Score} is built on \textit{AFG}, \textit{AFCG}, \textit{DocRE}, and \textit{LLM Validation}, we randomly select 30 records to identify the sources of errors. Upon manual inspection, we find that most errors occurred during \textit{DocRE} and \textit{LLM Validation}. This is because \textit{AFG} is fundamental for LLMs, and \textit{AFCG}, which involves NLI, is generally handled well by LLMs. However, for \textit{DocRE}, GPT-4 struggles to differentiate between sentence-level and document-level relations, leading to inaccuracies.
Additionally, the static threshold used for filtering document-level relations may be either too strict or too lenient, further contributing to errors. In \textit{LLM Validation}, we prompt the model to focus on newly added information, which requires a high level of reasoning capability of LLMs. To make \textsc{EVA-Score} more aligned with human judgments, we should either use more powerful LLMs for \textit{DocRE} or develop novel methods for relation extraction. The validation process can be improved through metrics like majority voting, which could enhance the reliability of the evaluations.
        
\subsection{Hyperparameter Tuning}

To mitigate redundancy, we prune DocRE triples that duplicate sentence-level facts by computing BERT-based cosine similarity and applying a single threshold~$\theta$. For each candidate $\theta$, we track three development-set metrics: \emph{Accuracy Rate}, \emph{Remaining Rate}, and \emph{Remaining Accuracy Rate} (Table~\ref{tab: threshold}). As $\theta$ increases, fewer triples are removed, which raises the Remaining Rate but typically lowers the Remaining Accuracy due to the admission of noisier near-duplicates. We select $\theta{=}0.65$ as a balanced operating point that preserves coverage while maintaining fidelity, yielding a cleaner evaluation set without sacrificing representativeness.

\begin{table}
    \centering
    \scalebox{0.9}{
    \begin{tabular}{l|c|c|c|c}
    \toprule
    Threshold & 0.60 & 0.65 & 0.70 & 0.75  \\
    \midrule
    Accuracy Rate & 0.823 & 0.861 & 0.886 & \textbf{0.920} \\
    \midrule
    Remaining Rate & 0.075 & 0.142 & 0.258 & \textbf{0.368} \\
    \midrule
    Remaining Accuracy Rate & \textbf{0.512} & 0.476 & 0.437 & 0.413 \\
    \bottomrule
    \end{tabular}  
    }
    \caption{Results of the performance of the filter under different thresholds. We use three metrics, \textit{Accuracy Rate}, \textit{Remaining Rate}, and \textit{Remaining Accuracy Rate}, for help. The largest value is denoted using \textbf{bold}.}
    \label{tab: threshold}
\end{table}

\section{CONCLUSION}
\label{conclusion}
We propose \textsc{EVA-Score}, an information-centric metric for abstractive LFS. The method extracts atomic facts, organizes them into logic chains, augments with document-level relations, and validates each information unit via LLMs. Experiments show higher correlation with human judgments than existing metrics, yielding an objective assessment. A re-evaluation of LLM summarizers highlights difficulties with long contexts and clarifies when existing metrics fail. Because \textsc{EVA-Score} quantifies information overlap, it may extend beyond summarization to other evaluation scenarios.


\newpage
\bibliographystyle{IEEEbib}
\bibliography{strings,refs}

\end{document}